\documentclass[conference]{IEEEtran}
\IEEEoverridecommandlockouts
\usepackage{cite}
\usepackage{amsmath,amssymb,amsfonts}
\usepackage{algorithmic}
\usepackage{graphicx}
\usepackage{textcomp}
\usepackage{xcolor}
\usepackage{makecell}
\def\BibTeX{{\rm B\kern-.05em{\sc i\kern-.025em b}\kern-.08em
    T\kern-.1667em\lower.7ex\hbox{E}\kern-.125emX}}

\usepackage{fancyhdr}
\begin{document}

\title{Interpretable Explainability in Facial Emotion Recognition and Gamification for Data Collection\\
}

\author{\IEEEauthorblockN{ Krist Shingjergji}
\IEEEauthorblockA{\textit{Educational Sciences} \\
\textit{Open University of the Netherlands}\\
Heerlen, The Netherlands \\
krist.shingjergji@ou.nl}
\and
\IEEEauthorblockN{Deniz Iren}
\IEEEauthorblockA{\textit{Center for Actionable Research} \\
\textit{Open University of the Netherlands}\\
Heerlen, The Netherlands \\
deniz.iren@ou.nl}
\and
\IEEEauthorblockN{Felix Böttger}
\IEEEauthorblockA{\textit{Center for Actionable Research } \\
\textit{Open University of the Netherlands}\\
Heerlen, The Netherlands\\
}
\and
\IEEEauthorblockN{Corrie Urlings}
\IEEEauthorblockA{\textit{Educational Sciences} \\
\textit{Open University of the Netherlands}\\
Heerlen, The Netherlands \\
}
\and
\IEEEauthorblockN{Roland Klemke}
\IEEEauthorblockA{\textit{Educational Sciences} \\
\textit{Open University of the Netherlands}\\
Heerlen, The Netherlands \\
}

}

\maketitle
\thispagestyle{fancy}

\begin{abstract}
Training facial emotion recognition models requires large sets of data and costly annotation processes. To alleviate this problem, we developed a gamified method of acquiring annotated facial emotion data without an explicit labeling effort by humans. The game, which we named Facegame, challenges the players to imitate a displayed image of a face that portrays a particular basic emotion. Every round played by the player creates new data that consists of a set of facial features and landmarks, already annotated with the emotion label of the target facial expression.  
Such an approach effectively creates a robust, sustainable, and continuous machine learning training process. We evaluated Facegame with an experiment that revealed several contributions to the field of affective computing. First, the gamified data collection approach allowed us to access a rich variation of facial expressions of each basic emotion due to the natural variations in the players' facial expressions and their expressive abilities. We report improved accuracy when the collected data were used to enrich well-known in-the-wild facial emotion datasets and consecutively used for training facial emotion recognition models. Second, the natural language prescription method used by the Facegame constitutes a novel approach for interpretable explainability that can be applied to any facial emotion recognition model. Finally, we observed significant improvements in the facial emotion perception and expression skills of the players through repeated game play.
\end{abstract}


\section{Introduction}

Facial expressions are imperative to non-verbal human communication as they provide a means of conveying information regarding the emotional state \cite{Scherer1986} as well as the behavioral intentions \cite{Fridlund} of the individual. Emotions are fundamental components of social interaction \cite{Matsumoto}, and the ability to express and perceive emotions is an invaluable asset for building social connections. The holy grail of affective computing is to empower computer systems with the ability to perceive and express emotions, and be able to form social ties with human users \cite{Picard2000}. Until very recently, this ability has been considered unique to humans. However, especially with the recent advances in Artificial Intelligence (AI), many studies have been conducted that focus on the automated recognition of emotions \cite{Maithri2022}. 

The common approach of training machine learning models for facial emotion recognition (FER) is supervised learning, which requires large sets of data \cite{Ko}. Specifically, deep FER models are challenged by a lack of sufficient data for training \cite{Li2020}. Collecting and curating such large datasets is a costly and time-consuming endeavour since labeling by human annotators is necessary \cite{Li2018}. This poses an obstacle to achieving significant performance improvements in emotion recognition research. 


Another major challenge lies in the explainability and interpretability of emotion recognition models. Studies mostly evaluate emotion recognition models using accuracy and confusion matrices; however, these metrics often fall short in reporting the utility of the models for humans. Interpretable models should provide explanations that are simple enough to be understood by their users, and are given in a language that is meaningful to them \cite{Gilpin}. The explainability of emotion recognition models has been very rarely addressed in literature. The approaches to achieve explainability are limited to \textit{model-agnostic} methods that explain the output of the model based on the inputs, and \textit{model-transparent methods} (e.g., \cite{Kumar}, \cite{Ghaleb}) that highlight the activation in different layers of artificial neural networks \cite{Jeyakumar}. However, neither approach necessarily provides human-friendly explanations that are interpretable by their users.

The challenges regarding collecting and curating excessive amounts of labeled data for training FER models, and yielding interpretable explanations from such models call for heterodox methods. In this study, we propose a gamification approach towards the collection of annotated facial emotion data. The proposed game, which is named; \textit{Facegame}, embodies a method for providing natural language prescription as feedback to the players, effectively serving as a means of achieving interpretable explainability. In summary, our contributions to the field of affective computing are as follows: 
\begin{itemize}
\item We present a gamification approach for rapidly collecting annotated facial emotion data that is rich in variety of facial expressions, in a low-cost, low-effort manner. 
\item We propose a novel approach for interpretable explanability by translating the intermediary facial features into natural language prescriptions, and providing them as an explanation for the emotion classifications provided by any FER model.
\item The presented gamification approach leads to significant improvements in the facial emotion perception and expression skills of the players.
\end{itemize}


The remaining of this paper is structured as follows. Section II provides a literature review on emotion recognition, explainable AI, and gamified data collection. Section III describes the core contributions. Section IV presents the details of our experimental study. Section V discloses the results of our experiments. Finally, Section VI provides a discussion on the theoretical and practical implications of our contributions, and concludes the paper. 


\section{Related Work}

\subsection{Face expressions, action units, and their automated recognition}

Facial expressions are a means for humans to express their emotions; thus, their automated detection is an important goal of affective computing. Facial expressions are movements and positions of the facial muscles that can be identified by Action Units (AUs); hierarchical components of movements of individual or groups of facial muscles that describe the changes in facial expressions \cite{EkmanFriesen}. There are studies focusing on the correlation between AUs and the basic emotions, namely, happiness, sadness, fear, disgust, anger, and surprise \cite{Ekman}. For example, Reisenzein et al. \cite{Reisenzein} reported coherence between amusement and smiling, and Wegrzyn et al. \cite{Wegrzyn} presented a detailed mapping between the basic emotions and different parts of the face, e.g., \textit{lid raiser} is essential for fear detection and \textit{lid tightener} for anger. Apart from the basic emotions, there are studies aiming at detecting more complex emotional states, such as confusion, by utilizing AUs \cite{Borges}.

The strong relationship between emotional facial expressions and AUs has motivated researchers to develop AU-detection algorithms as well as curating AU-labeled face expression datasets such as CK+ \cite{Lucey} and DISFA \cite{Mavadati}. For instance, Baltrušaitis et al. \cite{Baltrusaitis} presented an AU occurrence and intensity algorithm based on Histogram of Oriented Gradients (HOG); a method to describe an image by the distribution of intensity gradients or edge direction \cite{Dalal2005}), and geometrical features (e.g., shape and landmarks; detection and localization of certain characteristic points  on the face \cite{Celiktutan2013}). This work also highlights the positive impact of using various datasets to the generalizability of the model performance. Shao et al. \cite{Shao} presented a framework for detecting 10 AUs using the attention mechanism, i.e., finding the region-of-interest for each AU. Jacob and Stenger \cite{Jacob} outperformed their previous model by employing a correlation network based on a transformer-encoder architecture, to capture the relationship between different AUs for a wide range of expressions of emotions. Other prominent examples of architectures for AU detection are the JAA-Net \cite{JAAnet}, which uses high-level features of face alignment for AU detection, and DRML \cite{Zhao} which uses feed-forward functions to induce regions on the face that are important.

\subsection{Explainable and Interpretable AI}

As AI finds widespread application across many domains, the need for explainable AI (XAI) is rapidly growing as well. However, most explainability approaches do not target end-users, and their outcomes are not directly interpretable by humans. One way to address this issue is to improve the transparency of AI models. Model transparency focuses on explaining ``how the system made a decision'' \cite{Rosenfeld}. There are models that are transparent by design, e.g., decision trees, and others that are ``black box'' and require additional tools for explainability \cite{Guidotti}. In recent years, explanation tools have been designed to provide users insights on the decision-making process of a system. The study of Jeyakumar \cite{Jeyakumar} showed that the users prefer explanations by example in most of the cases. Rosenfeld \cite{RosenfeldXAI} presents a set of metrics that are suitable for evaluating the effectiveness of explainable AI, namely, i) the difference between the explanations' logic and the agent’s actual performance, ii) the number of rules in the agent’s explanation, iii) the number of features used to construct the explanation, and iv) the stability of the agent’s explanation.

\subsection{Crowdsourcing and Gamification for Data Collection}


Crowdsourcing is defined as the act of outsourcing a task that is commonly performed by designated agents to a large number of individuals \cite{Howe}. Crowdsourcing has been used both by industry and scientific communities for a variety of purposes, one of which is labeled data collection. In most cases, crowd workers complete a task with the motivation of monetary gain. Even though this approach has been proven cost-effective, it has also been criticized because it potentially leads to questionable data quality unless the necessary quality assurance mechanisms are put in place \cite{Iren}. A subtype of crowdsourcing, games-with-a-purpose \cite{Quinn}, provides a different kind of incentive \cite{Hosseini} for the workers to complete the tasks to the best of their abilities, and it generally incurs no additional costs to the employer. The design of crowdsourcing tasks in the form of a game is considered a part of a much larger concept; gamification. Gamification can be defined as a technique of using game elements in non-game systems to improve user experience and engagement \cite{Deterding}, increasing the motivation of the respondents by satisfying psychological and social needs \cite{Aparicio}. Gamification of data collection finds application in different domains \cite{Gurav}, such as education \cite{Lopez-Jimenez}\cite{Wang} and health \cite{Tobien}.

\section{Gamified Data Collection and Interpretable FER}
In this section, we present our solution that addresses the challenges of labeled data collection for FER model training and devising human-friendly explanations for emotion recognition systems. Specifically, we elaborate on the gamified data collection approach of Facegame and the underlying interpretable FER method. 

\subsection{Facegame}
Emotional facial expressions emerge rapidly and mostly involuntarily on human faces. Nevertheless, humans are exceptionally good at recognizing even the subtlest cues that appear on the faces of others. Even though humans inherently possess these abilities, it is surprisingly challenging to exercise them deliberately. The motivation of Facegame is to provide the players with a challenging way to exercise the skills of facial expression perception and mimicking. 

\begin{figure}[htbp]
\centerline{\includegraphics[width =0.47\textwidth]{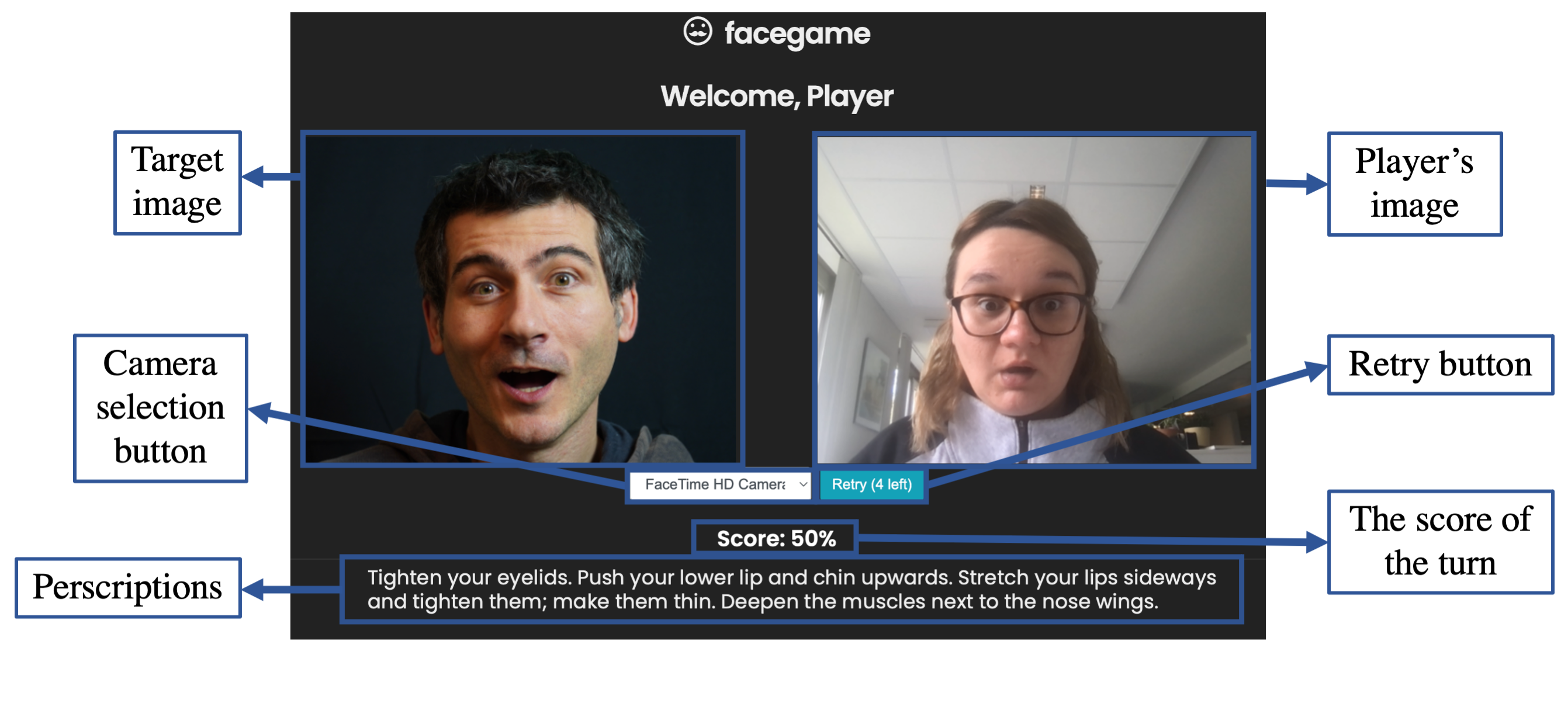}}
\caption{The interface of the Facegame.}
\label{fig:interface}
\end{figure}

In Facegame, the goal of the players is to mimic the facial expression shown on a target image. The interface of Facegame (Fig. \ref{fig:interface}) displays two images together; (a) a target image from the database of the game, which contains a face that exhibits a certain basic emotion and (b) the player’s camera feed.  Thus, the interface allows the player to compare both faces and try to imitate the target face.

\begin{figure}[htbp]
\centerline{\includegraphics[width =0.45\textwidth]{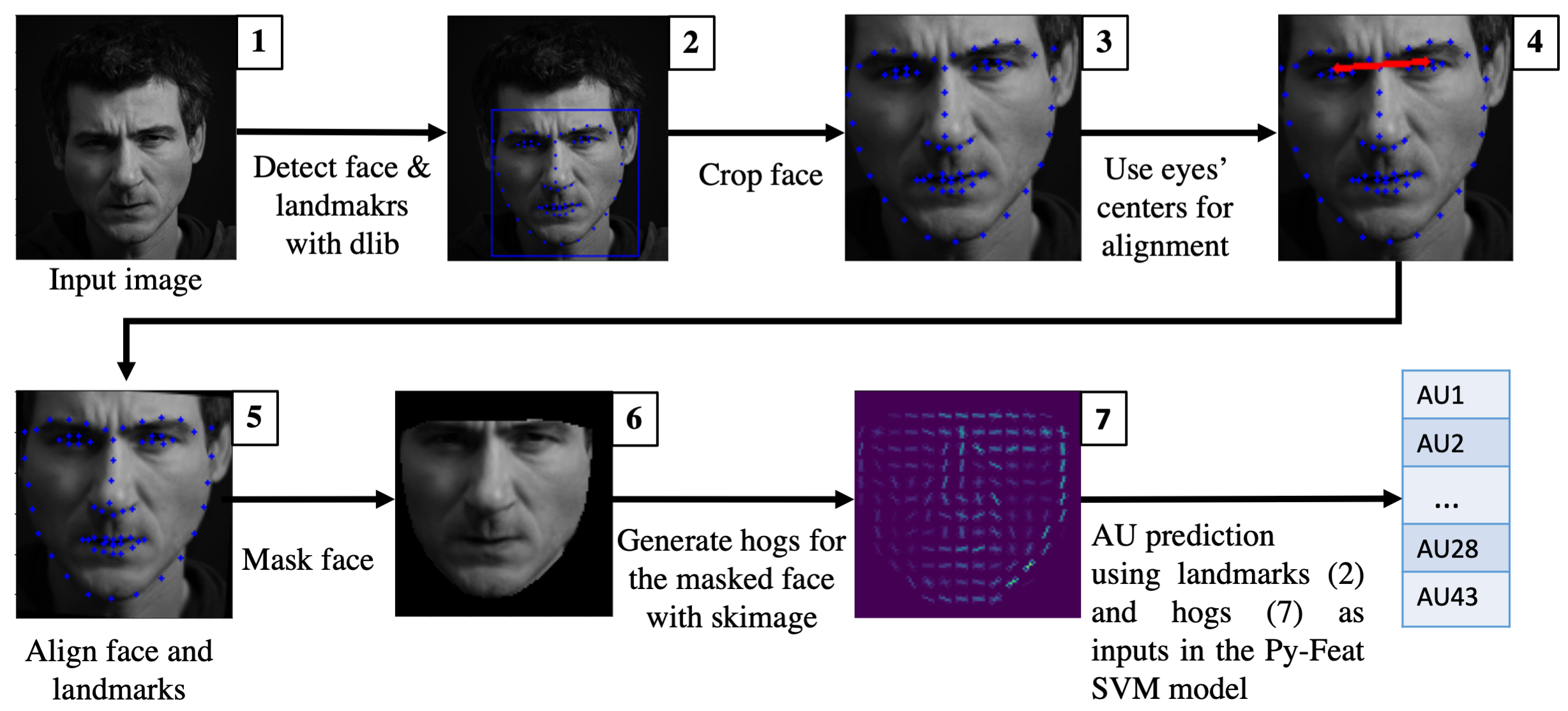}}
\caption{The pipeline of the AU detection.}
\label{fig:pipeline}
\end{figure}

All target images in the database are labeled based on six basic emotions \cite{EkmanFriesen} by human experts, as well as the 20 AUs (Table \ref{tab:AUs} ) automatically using a pre-trained classifier provided by Py-Feat \cite{Cheong}. 
The classifier's inputs are the following two vectors: the facial landmarks, a $(68 \times 2)$ vector of the landmark locations that is computed with the dlib package\cite{dlib}, and the HOGs; a vector of $(5408 \times 1)$ features that describe an image as a distribution of orientations \cite{Dalal2005}. The HOGs are calculated for the faces that are aligned using the position of the two eyes, and masked using the positions of the landmarks. The model's output is a list of AUs detected on the face image. The pipeline of the AU detection is shown in Fig. \ref{fig:pipeline}.

The player is given five seconds to mimic the target expression. Afterward, the player's image is processed and automatically labeled with AUs. The set of AUs on the target image are already known beforehand. Consecutively, the Jaccard Index of the two AU sets yields the score as seen in the Equation \ref{eq:score}; P and T depicting the AU sets of the player and target respectively. Players can retry imitating the same facial image to increase their scores or move on to another image.

\begin{equation}
Score(P,T)= \frac{|P\cap T|}{|P \cup T|}\label{eq:score}
\end{equation}

\subsection{Quality of data collected from the game}
\label{CNNs}

Every time a player plays the game, new data are generated. The score as a game element provides feedback to the player, and the players are motivated to do better and improve their scores, thus, generating better representations of the target facial emotion. This acts as an inherent quality assurance mechanism. Nevertheless, some players may show poor performance due to various reasons. For instance, they might be just exploring and testing the game, or their lighting conditions might be sub-optimal, or they simply do not feel motivated to do good in the game. In any case, such turns would yield low scores, and the data originating from players that consistently score low can easily be eliminated, thus, diminishing noise in data.

The turns that yield a high score are considered good representations of the emotional face expression on the target image, which is already labeled with one of six basic emotions. Thus, the players' image can automatically be annotated with the same label as the target image. The minor differences between the player and target AU sets provide a desirable variance in the distribution of AUs corresponding to a certain emotional face expression. This way, the variance in the AU distribution is created naturally by human players instead of automatically generated by means of simulation, potentially improving the in-the-wild performance of FER when used in training. 

\begin{table}[h!]
\begin{center}
                         
\begin{tabular}{|p{1.8cm}|p{1.8cm}|p{1.8cm}|p{1.8cm}|}
\hline
{\makecell[c]{ AU1 }} & {\makecell[c]{ AU2 }} & {\makecell[c]{ AU4 }} & {\makecell[c]{ AU5 }}   \\
\makecell[c]{\includegraphics[scale=0.2]{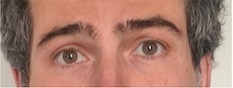}} & \makecell[c]{\includegraphics[scale=0.25]{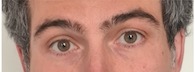}} & \makecell[c]{\includegraphics[scale=0.25]{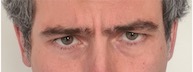}} & \makecell[c]{\includegraphics[scale=0.25]{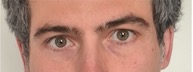}} \\
{\makecell[c]{Inner Brow \\ Raiser}} &
{\makecell[c]{Outer Brow \\ Raiser}} &
{\makecell[c]{Brow Lowerer}} &
{\makecell[c]{Upper Lid \\ Raiser}} \\ \hline \hline

{\makecell[c]{ AU6 }} & {\makecell[c]{ AU7 }} & {\makecell[c]{ AU9 }} & {\makecell[c]{ AU10 }}   \\
\makecell[c]{\includegraphics[scale=0.2]{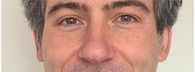}} & \makecell[c]{\includegraphics[scale=0.25]{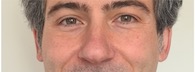}} & \makecell[c]{\includegraphics[scale=0.25]{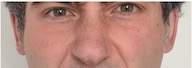}} & \makecell[c]{\includegraphics[scale=0.25]{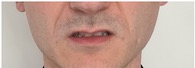}}\\
{\makecell[c]{Cheek Raiser}} &
{\makecell[c]{Lid Tightener}} &
{\makecell[c]{Nose Wrinkler}} &
{\makecell[c]{Upper Lip \\ Raiser}} \\ \hline \hline
{\makecell[c]{ AU11 }} & {\makecell[c]{ AU12 }} & {\makecell[c]{ AU14 }} & {\makecell[c]{ AU15 }}   \\
\makecell[c]{\includegraphics[scale=0.2]{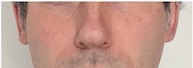}} & \makecell[c]{\includegraphics[scale=0.25]{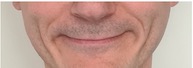}} & \makecell[c]{\includegraphics[scale=0.25]{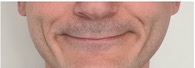}} & \makecell[c]{\includegraphics[scale=0.25]{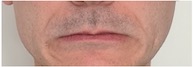}} \\
{\makecell[c]{Nasolabial \\ Deepener}} &
{\makecell[c]{Lip Corner \\ Puller }} &
{\makecell[c]{Dimpler}} &
{\makecell[c]{Lip Corner \\ Depressor}} \\ \hline \hline
{\makecell[c]{ AU17 }} & {\makecell[c]{ AU20 }} & {\makecell[c]{ AU23 }} & {\makecell[c]{ AU24 }}   \\
\makecell[c]{\includegraphics[scale=0.2]{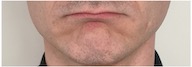}} & \makecell[c]{\includegraphics[scale=0.25]{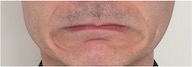}} & \makecell[c]{\includegraphics[scale=0.25]{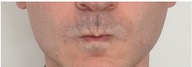}} & \makecell[c]{\includegraphics[scale=0.25]{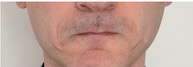}} \\
{\makecell[c]{Chin Raiser}} &
{\makecell[c]{Lip Stretcher }} &
{\makecell[c]{Lip Tightener}} &
{\makecell[c]{Lip Pressor}} \\ \hline \hline

{\makecell[c]{ AU25 }} & {\makecell[c]{ AU26 }} & {\makecell[c]{ AU28 }} & {\makecell[c]{ AU43 }} \\
\makecell[c]{\includegraphics[scale=0.2]{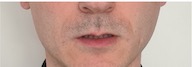}} & \makecell[c]{\includegraphics[scale=0.25]{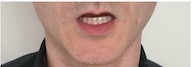}} & \makecell[c]{\includegraphics[scale=0.25]{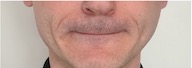}} & \makecell[c]{\includegraphics[scale=0.25]{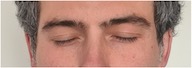}} \\
{\makecell[c]{Lips Part}} &
{\makecell[c]{Jaw Drop}} &
{\makecell[c]{Lip Suck}} &
{\makecell[c]{Eyes Closed}} \\ \hline 
\end{tabular}
\end{center}
\caption{The 20 AUs predicted by the AU detection model.}
\label{tab:AUs}
\end{table}

\subsection{Interpretable FER Explanation}
\begin{figure}[htbp]
\centerline{\includegraphics[scale=0.22]{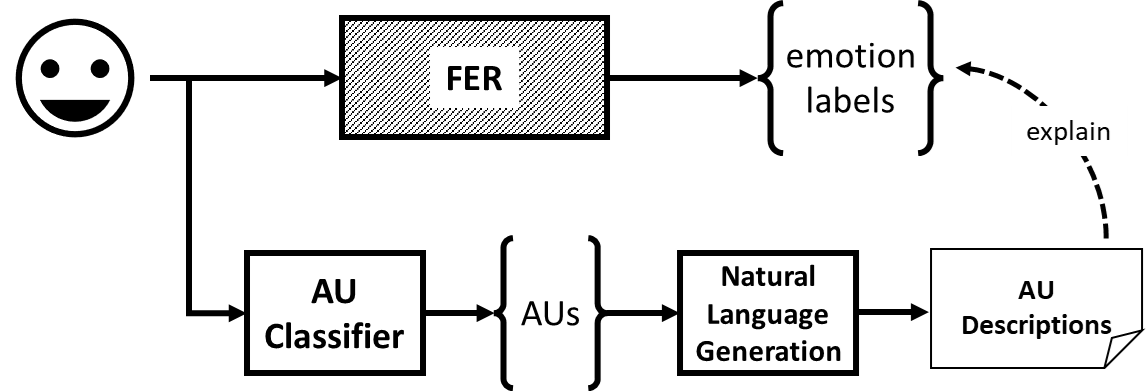}}
\caption{A graphical representation of the overall explainability model.}
\label{fig:overall_model}
\end{figure}

Even though there is no precise formula for how combinations of AUs translate into emotional expressions, some strong correlations exist. For instance, a happy face generally exhibits a smile, which is characterized by the existence of the ``lip corner puller.'' 
We propose using AUs as a means for explaining the the outcome of FER models. Specifically, we utilize AU detection parallel to FER, and translate the identified AUs into natural language descriptions, which constitute human-friendly, interpretable explanations of FER (Figure \ref{fig:overall_model}). The natural language descriptions are generated by a rule-based dictionary approach. 

We created the dictionary of AU descriptions based on the definitions on Facial Action Coding System \cite{EkmanFriesen}. The dictionary contains AU combinations categorized based on the facial muscle types and areas of appearance: cheeks, eyebrows, eyelids, lips, chin and nose, mouth, horizontal, oblique, and orbital. Every combination of the AUs that fall into these categories are represented in the dictionary. One example entry from the dictionary is as follows: 

\begin{quote}
Eyebrows, AU4: ``brow lowerer'' 

description: ``eyebrows are lowered.'' 

prescription (+): ``lower your eyebrows.''

prescription (-): ``do not lower your eyebrows.''
\end{quote}

The prescriptions of what players must do in order to improve their scores are given based on the outcome of a comparison between the target AU set and the player AU set. The intersection of both sets (\textit{P} for the player AUs and \textit{T}; target AUs) are the correctly mimicked AUs, while the difference between them show two kinds of mistakes; The set \textit{P-T} includes the AUs that should be removed from the player’s expression, and the set \textit{T-P} covers the AUs that are missing on the player’s expression to  mimic the target successfully. The AUs in both sets of mistakes are expressed as natural language prescriptions in different polarities; for example; ``raise your eyebrows'' and ``do not raise your eyebrows''.

\section{Methodology}

In this section, we present the overall methodology of this study, and describe the experimental setup for the data collection and analysis. In order to evaluate the efficacy and utility of the proposed gamified data collection method, i.e., Facegame, we pose the first research question as follows. 

\begin{quote}
\textit{RQ1: Do the data generated by the players represent emotional facial expressions accurately?}
\end{quote}

We also hypothesize that through playing the game repeatedly, players exercise deliberate practice of their face expression and perception skills, and thus, be able to improve them. Hence, our second research question is as follows. 

\begin{quote}
\textit{RQ2: Do the players improve their facial expression and perception skills through repeated play?}
\end{quote}

Finally, to evaluate how the proposed explanation method is perceived, and whether this yield an improved level of understanding regarding the outcome of the FER model. Hence, we formulate the third research question as follows.

\begin{quote}
\textit{RQ3: Do the natural language explanations help players understand the outcome of the FER model?}
\end{quote}

\subsection{Experiment setup and procedure}

In the experiment design, we adjusted the Facegame slightly. Participants were asked to play six rounds of the game, each round corresponding to one of the six target images, and each image representing one basic emotion. Participants were shown each target image five times in a row to be able observe the score change in each try. They were given five seconds, indicated by a countdown on the screen, to mimic the face. Following, the score of their attempt was displayed (Eq. \ref{eq:score}). 
To examine the potential effect of the natural language prescriptions, we divided the participants into two groups. One group received only the score (control group), and the other received the natural language prescriptions as well as the score (treatment group)

\subsection{Analysis of the collected emotional facial image data}

To answer the first research question, we analyzed the collected emotional facial image data, and evaluate to what extend the collected data can be used to model facial expressions of emotions. To this end, we used two commonly used in-the-wild data sets; Facial Expression Recognition 2013 (FER2013) \cite{FER2013}, and The Real-world Affective Face Database (RAF-DB) \cite{RAF1} \cite{RAF2}, and we trained a simple neural network classifier.  

FER2013 is a large-scale dataset that includes images collected automatically by the Google image search API labelled with the six basic emotions and neutral. RAF-DB is a real-world dataset that includes images of faces collected from the Internet and manually labelled with crowdsourcing. The neural network architecture we trained is similar to the baseline model used in the study of Bishay et al. \cite{CNNs}, a shallow convolutional neural network (CNN) including four  convolutional layers with 32,  32,  64,  and  64 filters  respectively and the ReLU activation function. The first three layers were followed by a max-pooling layer with a  $2 \times 2$ filter, and the last one by a flatten layer. The final two layers are fully-connected. The  first fully-connected layer has 96 neurons, while the second has six sigmoid units representing the predictions of the six emotions.

Considering that FER2013 and RAF-DB are large-scale, unbalanced datasets, samples were taken in order to respect the size and class distribution of the Facegame data. The samples were created by random sampling without replacement 200 and 50 instances from each of the six emotion from the training and testing set, respectively. With this technique we obtained a balanced sample of each set containing 1,200 instances for training and 300 instances for testing. To achieve generalizabile observations, we formed five different samples from each of both FER2013 and RAF-DB datasets. Finally, we compared the performance of the models trained on each sample set twice; first, without the inclusion of the Facegame Data, second, with the inclusion of Facegame Data.

\subsection{Participant Survey}
To receive quantitative and qualitative feedback regarding the design of Facegame, and to partially answer the third research question, we prepared a questionnaire. The participants were asked to fill in the questionnaire after completing the game experiment.
The questions covered demographics questions, i.e., age and gender, technical information, i.e., type of device and browser used for the game, and their quantitative and qualitative feedback on the game. The quantitative feedback was given with a 5-Likert scale score \textit{(Very Satisfied, Somewhat Satisfied, Neutral, Somewhat Unsatisfied, Very Unsatisfied)} on different aspects of the game; the ease of use, time to load and browser compatibility, and design. The survey for the participants in the treatment group included additional information regarding the natural language prescriptions. Specifically, they were asked to give a score on the usefulness and the design of the prescriptions. The qualitative feedback was sought with two open-ended questions about comments and suggestions regarding the functionality and the design of the game. Similar to the quantitative feedback, the participants of the treatment group were asked two additional open-ended questions regarding the clarity and other comments on the prescriptions that were displayed. The analysis of the survey data consisted of descriptive statistics on the age of the participants, quantitative analysis of the satisfaction scores regarding the design aspects of the game, and qualitative analysis on the open-ended questions.

\subsection{Analysis of Skill Improvement}

To answer the second and the third research questions, we analyzed the consecutive scores of the players in each round. Specifically, we compared the first score of each round against the mean of the remaining four scores of the same round. 
We tested the significance of the difference between the mean of the scores for both groups, and also between the groups using a t-test.  

\section{Results}
\subsection{Facegame Scores and Skill Improvement}

A total number of 36 individuals participated in the experiment of which 18 received the natural language prescriptions while the remaining 18 received only a score. In total, 216 games were played, each game yielding five consecutive scores: $S_1, S_2, S_3, S_4,$ and $S_5$. We examined the score change by comparing the distributions of S1 and the mean of the rest; $M_{rest}=M(S_2,S_3,S_4,S_5)$.

Our results show that, for all games $(N=216)$ the score increased significantly between $S_1$ $(M=0.4$, $SD=0.23)$ and $M_{rest}$ $(M=0.45$, $SD=0.21)$ with $t_{(215)}=2.61$, $p<0.01$ which is a clear indication of the learning effect of Facegame. 

For the control group $(N=108)$, the same comparison also yielded a significant increase in the distributions $S_1 (M=0.41, SD=0.23)$ and $M_{rest} (M=0.46, SD=0.21)$,  $t_{(107)} = 2.26$, $p = 0.02.$. For the treatment group $(N=108)$ that received natural language prescriptions, the comparison showed an insignificant increase in the score  distributions $S_1 (M=0.40, SD=0.24)$ and $M_{rest} (M=0.44, SD=0.22)$,  $t_{(107)} = 1.44$, $p = 0.15.$

Additionally, we investigated the number of times a game ends with an increased score for both groups. Our observations showed that $62.9\%$ of the games resulted with an increased score when the participants received natural language descriptions, while the score increased $57.4\%$ when the participants received only a score. 

\subsection{Survey Results and Player Feedback}

\begin{figure}
  \includegraphics[width =0.5\textwidth]{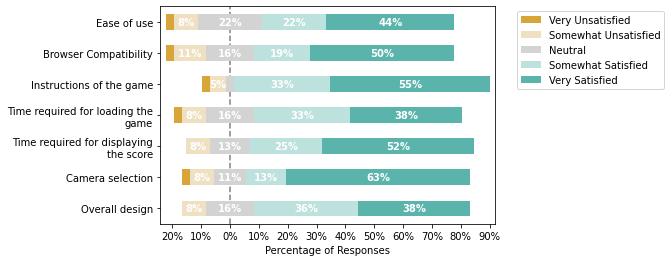}
  \caption{Likert scale}
  \label{fig:likert_overall}
\end{figure}

\begin{figure}
  \includegraphics[width =0.5\textwidth]{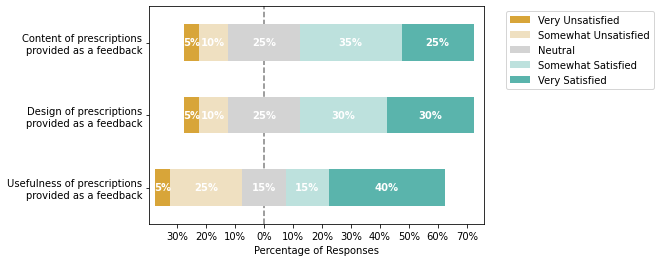}
  \caption{Likert scale}
  \label{fig:likert_T}
\end{figure}

In the online survey, 36 participants (22 male; 14 female) provided feedback. The age of the participants ranged between 25 and 55 years ($M=33.77$, $SD=7.66)$. Fig. \ref{fig:likert_overall} presents the satisfaction scores provided by all participants regarding various aspects of the design of the game. 
Fig. \ref{fig:likert_T} shows the feedback of the participants who received natural language prescriptions ($N=18$) in the experiment regarding their content, design, and usefulness.
The results revealed that most of the participants were satisfied or neutral to the content and the design of the prescriptions while attention should be given to the usefulness prescriptions.

We manually semantically grouped the answers from the two open-ended questions. The open-ended question that inquires participants’ suggestions on the prescriptions shows that participants find the use of visualization, prioritization, and personalization useful. Specifically, four participants mentioned that on-screen visualization of the part of the face that they needed to change would help them follow the natural language prescriptions better. Three of the participants found the text too long to read in a short time span and suggested the display of a few of the most important instead. Two participants indicated that more personalized prescriptions would be helpful. Lastly, the 12 participants that commented on the natural language of the prescriptions stated that they found them understandable.

\subsection{Facial Emotion Recognition with Facegame Data}
\begin{figure}[htbp]
\centerline{\includegraphics[width =0.45\textwidth]{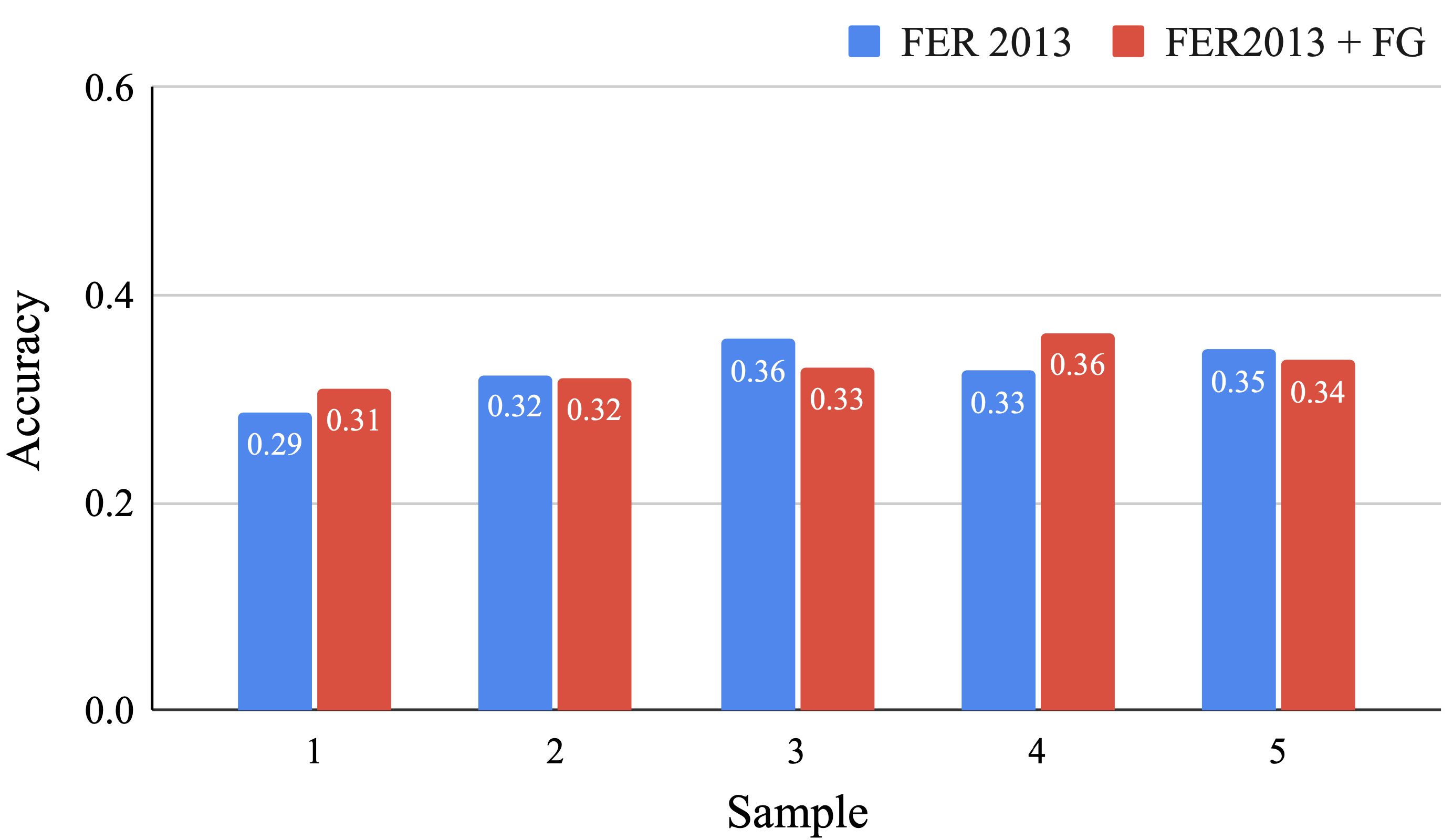}}
\caption{Performance of the models trained on FER2013 dataset and FER2013 dataset combined with the Facegame Dataset (FG)  }
\label{fig:accuracies fer}
\end{figure}

\begin{figure}[htbp]
\centerline{\includegraphics[width =0.45\textwidth]{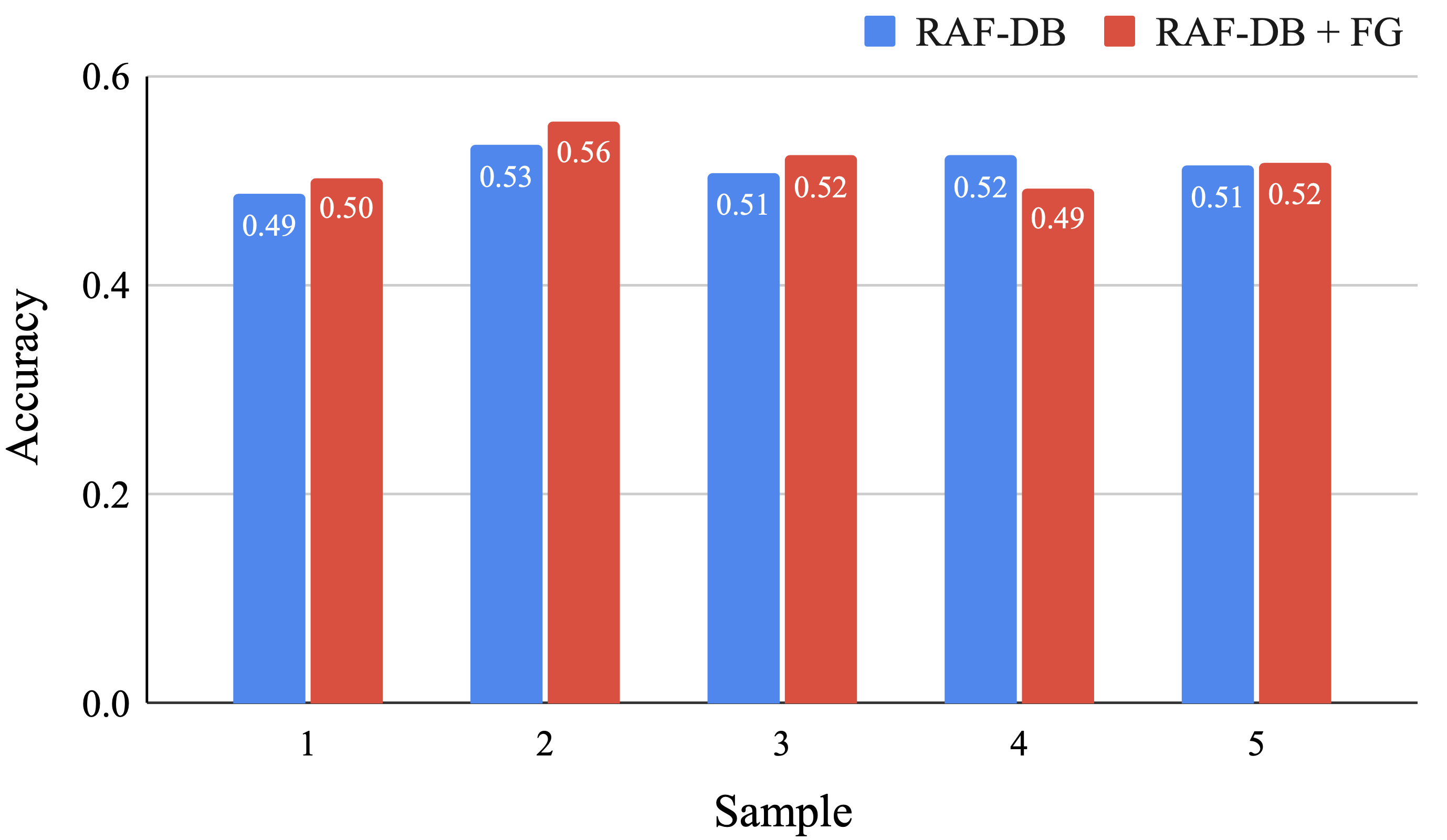}}
\caption{Performance of the models trained on RAF-DB and RAF-DB combined with the Facegame Dataset (FG)}
\label{fig:accuracies_raf}
\end{figure}

The Facegame Data that were collected in the experiment consist of 636 images depicting the six basic emotions (86 angry, 89 disgusted, 127 fearful, 132 happy, 99 sad, and 103 surprised). The accuracy for each of the five samples for FER2013 and RAF-DB is shown in Fig. \ref{fig:accuracies fer} and Fig. \ref{fig:accuracies_raf}, respectively.
The average accuracy of the model trained on instances from FER2013 is 32.80\%, while for the model trained on the instances from the combination of FER2013 and the Facegame Data is 33,20\%. Similarly, the average accuracy of the model trained on instances from RAF-DB is 51,27\%, while for the model trained on the instances from the combination of FER2013 and the Facegame Data is 51,87\%. The observed low accuracies of the models were expected considering the nature of the sets and the simplicity of our neural network architecture. The images in FER2013 and RAF-DB represent in-the-wild conditions with subtle facial expressions. Therefore, the emotion recognition task is more challenging, and it requires complex classifiers to achieve better accuracy. Moreover, our goal was to examine the quality of the data collected via Facegame in comparison with the existing in-the-wild sets, and for this purpose, a simple neural network sufficed. The results show the we were able to collect labelled data that can potentially improve FER in-the-wild.

\subsection{The Mapping of AUs to Emotion Classes and the Variability}
\begin{figure}[htbp]
\centerline{\includegraphics[width =0.62\textwidth]{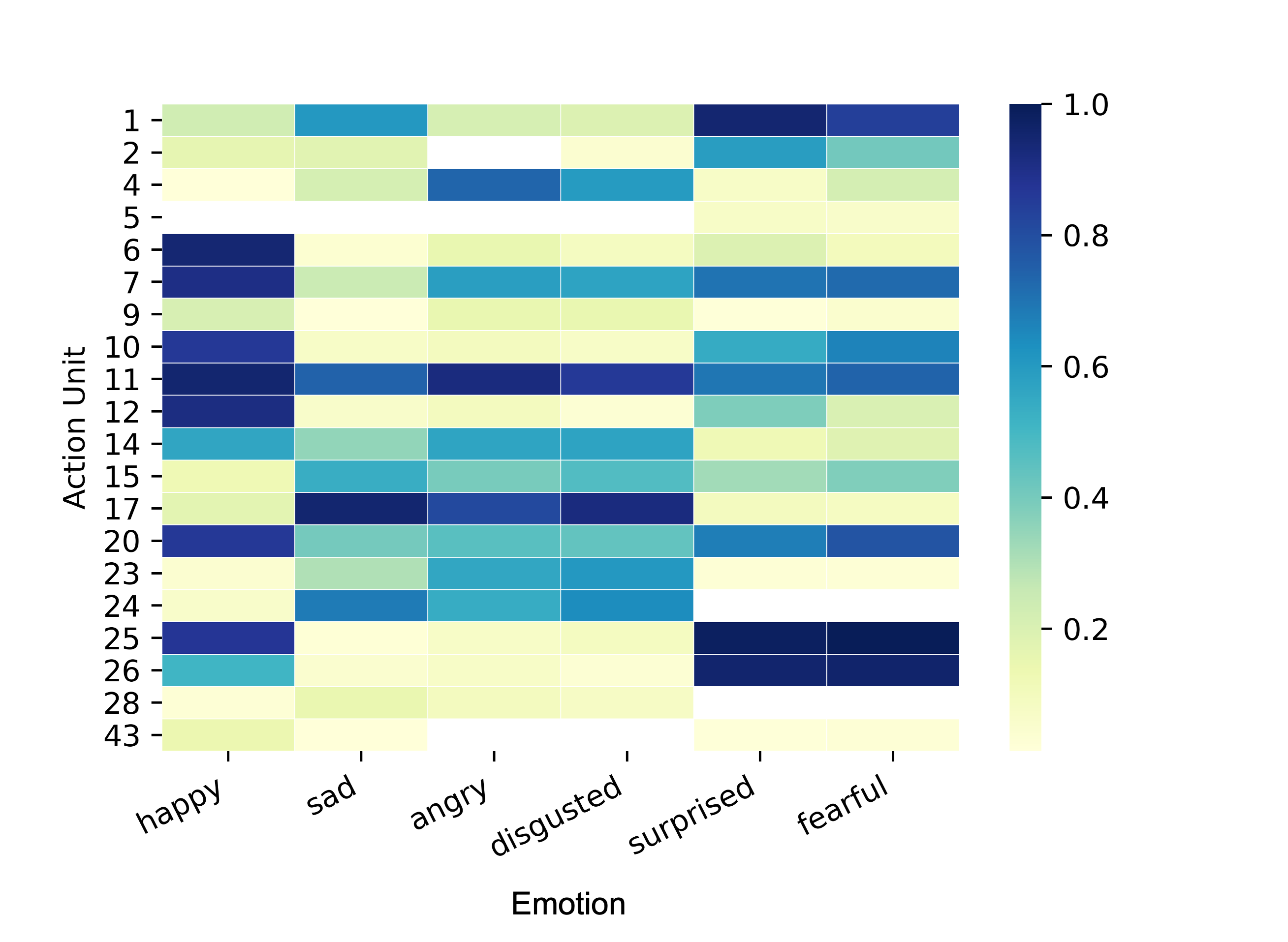}}
\caption{Heatmap of the occurrences of AUs detected on the participants and the emotions of the targeted image (threshold = $\frac{1}{3}$).}
\label{fig:frequences}
\end{figure}

The results of the correlation between the six basic emotions and the AUs are shown in Fig. \ref{fig:frequences}. For this analysis, the data from the trials that scored below $\frac{1}{3}$ were excluded. The results suggest that we were able to capture some strong correlations between certain emotional facial expressions and their signature AUs. For instance, lips part (AU25) and jaw drop (AU26) highly correlate with both surprise and fear, while lip corner puller (AU12) highly correlates with happiness. The high occurrence number of nasolabial deepener (AU11) is an indication of the underlying Py-Feat model creating a high false-positive rate for AU11, which needs further examination. The results show that we were able to define the emotion classes as a distribution of multiple AUs. Such naturally occurring variety in facial expression data can potentially be used to improve FER in-the-wild.

\section{Discussion and Conclusion}

In this paper, we introduced a novel approach for explainable FER that promises three related contributions. First, by means of gamification, we have developed a method for collecting annotated face expression data continuously, which allows us to describe the facial expressions of six basic emotions as a distribution of AUs. Secondly, we proposed and evaluated an interpretable FER explainability method that uses AUs as features to describe the outcomes of FER models, i.e., facial emotion classes. The experimental observations indicate that the natural language explanations of face expressions are interpretable by humans. Our quantitative and qualitative results highlight improvement opportunities regarding the design of Facegame and how we communicate the face expression explanations. Finally, we observed that the players were able to improve their face expression and perception skills by playing the game. 

Our results have potential theoretical and practical implications. Our method of acquiring nuanced face expressions (i.e., distributions of AUs) that correlate with facial emotion classes provides a means to improve the performance of in-the-wild FER models. Such performance improvements may pave the way for novel practical approaches in many domains, such as online synchronous learning \cite{Shingjergji}. Moreover, the gamification approach offers a sustainable, continuous self-training process of FER models. Our explainability method that uses AUs as intermediary features to describe facial emotions provides a novel approach towards achieving interpretable, human-friendly explanations of FER models. In this study, we conducted our experiment and analyses on a limited sample. In the future, based on the observations and feedback collected during this study, we will continue this line of research and improve the ways of providing prescriptions to the players of Facegame by combining the natural language explanations with graphical methods. Finally, the quality of the data was assured by using a threshold of the score as a selection criterion. The score of the trial was calculated by the presence/absence of the AUs regardless of the intensity which has the risk of including data with exaggerating facial expressions and introducing bias to the data set \cite{overexaggerating}. In future studies, we aim to include the intensity of the AUs as well as a higher threshold in the selection criteria to increase the accuracy and fairness of the collected data set.  

The goal of this study is limited to the evaluation of gamification as a data collection method. In the future, we aim to collect more data by crowdsourcing a large number of players which will result in a more in-depth analysis of the collected data set. In this setting, the player was asked to mimic the facial expressions of a face displaying one emotion and not triggered to experience that emotion. This approach has limitations considering that there are differences between the posed and spontaneous facial expressions in morphological and dynamic aspects of certain emotions \cite{posedvsspontaneous} as well as demographical mismatches between the target image and the player. In future studies, we aim to increase the variance of the dataset by including in the set of the target images faces with spontaneous facial expressions as well as increasing the demographic diversity of the target images.

\section*{Ethical Impact Statement}
In this study, the personal data collection was limited to facial images of the players. The collected data were securely stored in the server of the research institute which can only be accessed by the researchers of this study, The participants were informed regarding the experiment and data collection, and they provided consent prior to taking part in the experiment. 



\begin{thebibliography}{00}

\bibitem{Scherer1986}K. R. Scherer, and P. H. Tannenbaum, ``Emotional experiences in everyday life: A survey approach,'' Motivation and Emotion, vol. 10, pp. 295--314, 1986

\bibitem{Fridlund} A.J. Fridlund, ``Human facial expression: An evolutionary view,'' Academic press, March 2014.

\bibitem{Matsumoto} D. Matsumoto, K. Dacher, M. N. Shiota, M. O'Sullivan, and M. Frank, ``Facial expressions of emotion,''The Guilford Press, 2008.

\bibitem{Picard2000} R. W. Picard, ``Affective computing,'' MIT press, July 2000.

\bibitem{Maithri2022}M. Maithri, U. Raghavendra, A. Gudigar, J. Samanth, D. B. Prabal, M. Murugappan, et al. ``Automated Emotion Recognition: Current Trends and Future Perspectives,'' Computer Methods and Programs in Biomedicine, pp. 106646, 2022.

\bibitem{Ko} B. C. Ko, ``A brief review of facial emotion recognition based on visual information,'' sensors, vol. 18, pp.401, February 2018.

\bibitem{Li2020} S. Li, and D. Weihong, ``Deep facial expression recognition: A survey,'' IEEE transactions on affective computing, 2020.

\bibitem{Li2018} S. Li, and D. Weihong, ``Reliable crowdsourcing and deep locality-preserving learning for unconstrained facial expression recognition,'' IEEE Transactions on Image Processing, vol. 28, pp.356--370, 2018.



\bibitem{Gilpin} L.H. Gilpin, D. Bau, B.Z. Yuan BZ, A. Bajwa, M. Specter, and L. Kagal. ``Explaining explanations: An overview of interpretability of machine learning,'' 2018 IEEE 5th International Conference on Data Science and Advanced Analytics (DSAA), pp. 80-89, October 2018.

\bibitem{Kumar} P. Kumar P, Kaushik V, and B. Raman. ``Towards the Explainability of Multimodal Speech Emotion Recognition,'' Interspeech, pp 1748--1752, 2021

\bibitem{Ghaleb} E. Ghaleb, A. Mertens, S. Asteriadis, and G. Weiss, ``Skeleton-Based Explainable Bodily Expressed Emotion Recognition Through Graph Convolutional Networks,'' 2021 16th IEEE International Conference on Automatic Face and Gesture Recognition (FG 2021), pp. 1--8, 2021.

\bibitem{Jeyakumar} J.V. Jeyakumar, J. Noor, Y.H Cheng, L. Garcia, and M. Srivastava, ``How can i explain this to you? an empirical study of deep neural network explanation methods,'' Advances in Neural Information Processing Systems, vol. 33, pp. 4211--4222, 2020.

\bibitem{EkmanFriesen}P. Ekman, and W. Friesen, ``Facial action coding system,'' Environmental Psychology \& Nonverbal Behavior, 1978.

\bibitem{Ekman} P. Ekman, ``Facial expression and emotion,'' American psychologist vol. 48, pp. 384, April 1993.

\bibitem{Reisenzein}R. Reisenzein, M. Studtmann, and G. Horstmann. ``Coherence between emotion and facial expression: Evidence from laboratory experiments,'' Emotion Review, vol. 5, pp. 16--23, January 2013.

\bibitem{Wegrzyn}M. Wegrzyn, M. Vogt, B. Kireclioglu, J. Schneider, and J. Kissler, ``Mapping the emotional face. How individual face parts contribute to successful emotion recognition,'' PloS one, vol. 12, pp. 11--12, May 2017.

\bibitem{Borges} N. Borges, L. Lindblom, B. Clarke, A. Gander, and R. Lowe, ``Classifying confusion: autodetection of communicative misunderstandings using facial action units,'' 2019 8th International Conference on Affective Computing and Intelligent Interaction Workshops and Demos (ACIIW), pp. 401--406, September 2019.

\bibitem{Lucey}P. Lucey, J. F. Cohn, T. Kanade, J. Saragih, Z. Ambadar, and I. Matthews, ``The extended cohn-kanade dataset (CK+): A complete dataset for action unit and emotion-specified expression,'' 2010 IEEE computer society conference on computer vision and pattern recognition-workshops, pp. 94--101, June 2010.

\bibitem{Mavadati}S.M. Mavadati, M.H Mahoor, K. Bartlett, P. Trinh, and J.F. Cohn, ``Disfa: A spontaneous facial action intensity database,'' IEEE Transactions on Affective Computing, pp. 151--160 2013 Mar 7;4(2):151--60.

\bibitem{Baltrusaitis}T. Baltrušaitis, M. Mahmoud, and P. Robinson, ``Cross-dataset learning and person-specific normalisation for automatic action unit detection,'' 2015 11th IEEE International Conference and Workshops on Automatic Face and Gesture Recognition (FG), vol. 6, pp. 1--6, May 2015.

\bibitem{Dalal2005}N. Dalal, and B. Triggs, ``Histograms of oriented gradients for human detection,'' In 2005 IEEE computer society conference on computer vision and pattern recognition (CVPR'05), vol. 1, pp. 886--893, June 2005.

\bibitem{Celiktutan2013} O. Çeliktutan, S. Ulukaya, and B. Sankur, ``A comparative study of face landmarking techniques,'' EURASIP Journal on Image and Video Processing, pp. 1--27, December 2013. 

\bibitem{Shao}Z. Shao, Z. Liu, J. Cai, Y. Wu, and L, Ma, ``Facial action unit detection using attention and relation learning,'' IEEE transactions on affective computing, October 2019.

\bibitem{Jacob}G.M. Jacob, and B. Stenger, ``Facial action unit detection with transformers,'' Proceedings of the IEEE/CVF Conference on Computer Vision and Pattern Recognition, pp. 7680--7689, 2021.

\bibitem{JAAnet}Z. Shao, Z. Liu, J. Cai, Y. Wu, and L, Ma, ``JAA-Net: joint facial action unit detection and face alignment via adaptive attention,'' International Journal of Computer Vision, vol. 129, pp. 321--340, February 2021.

\bibitem{Zhao}K. Zhao, W.S Chu, and Zhang, ``Deep region and multi-label learning for facial action unit detection,'' Proceedings of the IEEE conference on computer vision and pattern recognition, pp. 3391--3399, 2016.

\bibitem{Rosenfeld}A. Rosenfeld, and A. Richardson, ``Explainability in human–agent systems,'' Autonomous Agents and Multi-Agent Systems, vol. 33, pp. 673--705 November 2019.

\bibitem{Guidotti}R. Guidotti, A. Monreale, S. Ruggieri, F. Turini, F. Giannotti, and D. Pedreschi, ``A survey of methods for explaining black box models,'' ACM computing surveys (CSUR). 2018 Aug 22;51(5):1-42.

\bibitem{RosenfeldXAI}A. Rosenfeld, ``Better metrics for evaluating explainable artificial intelligence,'' Proceedings of the 20th international conference on autonomous agents and multiagent systems, pp. 45--55, May 2021.

\bibitem{Howe}J. Howe J, ``The rise of crowdsourcing,'' Wired magazine, vol. 6, pp. 1--4, June 2006.

\bibitem{Iren}D. Iren, S. Bilgen, ``Cost of quality in crowdsourcing,'' Human Computation, vol. 1, December 2014.

\bibitem{Quinn}A.J Quinn, and B.B Bederson, ``Human computation: a survey and taxonomy of a growing field,'' Proceedings of the SIGCHI conference on human factors in computing systems, pp. 1403--1412, May 2011.

\bibitem{Hosseini} M. Hosseini, A. Shahri, K. Phalp, J. Taylor, and R. Ali, ``Crowdsourcing: A taxonomy and systematic mapping study,'' Computer Science Review, vol. 17, pp. 43--69, August 2015.

\bibitem{Deterding}S. Deterding, D. Dixon, R. Khaled, and L. Nacke, ``From game design elements to gamefulness: defining" gamification,'' Proceedings of the 15th international academic MindTrek conference: Envisioning future media environments, pp. 9--15, September 2011

\bibitem{Aparicio}A.F. Aparicio, F.L. Vela, J.L. Sánchez, and J.L Montes, ``Analysis and application of gamification,'' Proceedings of the 13th International Conference on Interacción Persona-Ordenador, pp. 1--2, October 2012.

\bibitem{Gurav}V. Gurav, M. Parkar, and P. Kharwar, ``Accessible and Ethical Data Annotation with the Application of Gamification,'' International Conference on Recent Developments in Science, Engineering and Technology, pp. 68--78, November 2019.

\bibitem{Lopez-Jimenez}J.J López-Jiménez, J.L. Fernández-Alemán, L.L González, O.G Sequeros, B.M Valle, J.A García-Berná, et al. ``Taking the pulse of a classroom with a gamified audience response system,'' Computer Methods and Programs in Biomedicine, vol. 213, pp. 106459, January 2022. 

\bibitem{Wang} A.I. Wang, and R. Tahir,  ``The effect of using Kahoot! for learning–A literature review'' Computers \& Education, vol. 149, pp. 103818, May 2020.

\bibitem{Tobien}P. Tobien, L. Lischke, M. Hirsch, R. Krüger, P. Lukowicz, A. Schmidt, ``Engaging people to participate in data collection,'' Proceedings of the 2016 ACM International Joint Conference on Pervasive and Ubiquitous Computing: Adjunct, pp. 209-212, September 2016.

\bibitem{Cheong}J.H. Cheong, T. Xie, S. Byrne, and L.J. Chang, ``Py-feat: Python facial expression analysis toolbox,'' arXiv preprint arXiv:2104.03509, April 2021.

\bibitem{dlib}D. E. King, ``Dlib-ml: A machine learning toolkit,'' The Journal of Machine Learning Research, vol. 10, pp. 1755--1758, 2009.


\bibitem{FER2013}I .J. Goodfellow, D. Erhan, P. L. Carrier, A. Courville, M. Mirza, B. Hamner et al, ``Challenges in representation learning: A report on three machine learning contests,'' International conference on neural information processing, Springer, Berlin, pp. 117--124, November 2013.

\bibitem{RAF1}S. Li, W. Deng, and J. Du, ``Reliable crowdsourcing and deep locality-preserving learning for expression recognition in the wild,'' IEEE Conference on Computer Vision and Pattern Recognition (CVPR), pp. 2852--2861, 2017.

\bibitem{RAF2}S. Li, and W. Deng, ``Reliable Crowdsourcing and Deep Locality-Preserving Learning for Unconstrained Facial Expression Recognition,'' IEEE Transactions on Image Processing, vol. 28, pp. 356--370, 2019.

\bibitem{CNNs}M. Bishay, A. Ghoneim, M. Ashraf, M. Mavadati, ``Which CNNs and Training Settings to Choose for Action Unit Detection? A Study Based on a Large-Scale Dataset,'' 16th IEEE International Conference on Automatic Face and Gesture Recognition (FG 2021), pp. 1--5, 2021.

\bibitem{Shingjergji}K. Shingjergji, D. Iren, C. Urlings, and R. Klemke, ``Sense the classroom: AI-supported synchronous online education for a resilient new normal,'' EC-TEL (Doctoral Consortium), pp. 64--70, January 2021.

\bibitem{overexaggerating}M. Mäkäräinen, J. Kätsyri, and T. Takala, ``Exaggerating facial expressions: A way to intensify emotion or a way to the uncanny valley?,'' Cognitive Computation, vol. 6. pp. 708--721, Decembe 2014.

\bibitem{posedvsspontaneous}S. Namba, S. Makihara, R.S. Kabir, M. Miyatani, and T. Nakao, ``Spontaneous Facial Expressions Are Different from Posed Facial Expressions: Morphological Properties and Dynamic Sequences,'' Current Psychology, vol 36, pp. 593--605, 2017

\end{thebibliography}
\end{document}